\documentclass[10pt,twocolumn,letterpaper]{article}

\usepackage[pagenumbers]{src/cvpr} %

\newcommand{\cmpart}{\textcolor{checkok}{(\checkmark)}}
\newcommand{\figref}[1]{Fig.~\ref{#1}}
\newcommand{\secref}[1]{Sec.~\ref{#1}}

\newcommand{\tabref}[1]{Table~\ref{#1}}

\makeatletter
\DeclareRobustCommand\onedot{\futurelet\@let@token\@onedot}
\def\@onedot{\ifx\@let@token.\else.\null\fi\xspace}
\def\eg{e.g\onedot}

\makeatother

\newcommand{\boldparagraph}[1]{\vspace{0.0cm}\noindent{\bf #1} }

\definecolor{darkgreen}{rgb}{0,0.7,0}

\definecolor{ellisred}{rgb}{0.87,0.44,0.38} %
\definecolor{ellisgreen}{rgb}{0.69,0.90,0.52} %
\definecolor{elliscyan}{rgb}{0.29,0.77,0.74} %
\definecolor{ellisorange}{rgb}{0.89,0.55,0.28} %
\definecolor{ellisblue}{rgb}{0.41,0.61,0.86} %
\definecolor{ellisviolett}{rgb}{0.57,0.34,0.51} %
\definecolor{darkellisgreen}{rgb}{0.5176470588235295, 0.6745098039215687, 0.38823529411764707} 
\definecolor{textviolet}{HTML}{804c70}
\definecolor{lightgray}{HTML}{E9E9E9}

\definecolor{waymogreen}{HTML}{00A74E}
\definecolor{tabpurple}{HTML}{9D62AD}
\definecolor{paigray}{HTML}{5C5C5C}
\definecolor{nurecgold}{HTML}{D4A72C}

\definecolor{amber}{HTML}{FFBE28}
\definecolor{flame}{HTML}{E65100}
\definecolor{indigo}{HTML}{1A237E}
\definecolor{cyan}{HTML}{00ACC1}

\definecolor{teal}{HTML}{006064}
\definecolor{fern}{HTML}{81C784}
\definecolor{mint}{HTML}{E8F5E9}
\definecolor{slate}{HTML}{263238}

\definecolor{camorange}{HTML}{F28E2B}
\definecolor{camcyan}{HTML}{76B7B2}
\definecolor{camgreen}{HTML}{59A14F}
\definecolor{cammauve}{HTML}{B07AA1}
\definecolor{cammustard}{HTML}{EDC948}
\definecolor{cambrown}{HTML}{9C755F}
\definecolor{camviolet}{HTML}{9467BD}
\definecolor{cammagenta}{HTML}{FF9DA7}
\definecolor{camolive}{HTML}{BCBD22}
\definecolor{camteal}{HTML}{275832}

\usepackage{times}
\usepackage{epsfig}
\usepackage{graphicx}
\usepackage{amsmath,amsfonts,amssymb,amsthm}
\usepackage{color}
\usepackage{caption}
\usepackage{subcaption}
\usepackage{xspace}
\usepackage{array}
\usepackage{cite}
\usepackage{booktabs}
\usepackage{multicol, blindtext}
\usepackage{xcolor}
\usepackage[labelfont={bf}]{caption}
\usepackage{pifont}%
\usepackage{minted}
\usepackage{xcolor}
\usepackage{tcolorbox}
\tcbuselibrary{minted, skins}
\usepackage{multirow}
\usepackage{graphicx}
\usepackage{subcaption}
\usepackage{realboxes}
\usepackage[table]{xcolor} %
\usepackage{diagbox} %
\usepackage{slashbox} %
\usepackage{wrapfig}
\usepackage{colortbl}
\usepackage{wrapfig}
\usepackage{rotating}

\newcommand{\xmark}{\ding{55}}
\newtcbox{\textcode}{on line, 
    boxsep=0pt, 
    left=2pt, 
    right=2pt, 
    top=1pt, 
    bottom=1pt,
    arc=2pt,        %
    auto outer arc,
    colback=lightgray, 
    colframe=lightgray, %
    fontupper=\ttfamily\color{black},
    boxrule=0pt
}
\definecolor{checkok}{HTML}{2E7D32}    %
\definecolor{xmarkno}{HTML}{B71C1C}    %
\newcommand{\cmgreen}{\textcolor{checkok}{\checkmark}}
\newcommand{\xmred}{\textcolor{xmarkno}{\xmark}}

\definecolor{cvprblue}{rgb}{0.21,0.49,0.74}
\usepackage[pagebackref,breaklinks,colorlinks,allcolors=cvprblue]{hyperref}

\def\confName{CVPR}
\def\confYear{2027}

\title{123D: Unifying Multi-Modal Autonomous Driving Data at Scale}

\author{
{Daniel Dauner}$^{1,2*}$ \quad
{Long Nguyen}$^{1,2,3*}$ \quad
{Valentin Charraut}$^{4}$ \quad
{Bastian Berle}$^{2}$ \quad
{Tianyu Li}$^{5}$ \\
{Jiabao Wang}$^{6}$  \quad
{Changhui Jing}$^{5}$  \quad 
{Maximilian Igl}$^{7}$ \quad
{Holger Caesar}$^{8}$ \quad 
{Boris Ivanovic}$^{7}$ \\
{Yiyi Liao}$^{6}$ \quad 
{Andreas Geiger}$^{1,2}$ \quad
{Kashyap Chitta}$^{1,3}$
\vspace{0.4cm}
\\
{$^{1}$}KE:SAI \quad
{$^{2}$}University of T{\"u}bingen, T{\"u}bingen AI Center \quad
{$^{3}$}ELLIS Institute T{\"u}bingen \\
{$^{4}$}Valeo Brain \quad
{$^{5}$}OpenDriveLab at Shanghai Innovation Institute \quad
{$^{6}$}Zhejiang University \\
{$^{7}$}NVIDIA Research  \quad
{$^{8}$}Delft University of Technology \\
}

\begin{document}

\maketitle
\def\thefootnote{*}\footnotetext{Equal contribution.}
\def\thefootnote{\arabic{footnote}}

\begin{abstract}
The pursuit of autonomous driving has produced one of the richest sensor data collections in all of robotics. However, public datasets are fragmented, as each release adopts different modalities, such as cameras, lidar, ego states, annotations, and HD maps, recorded at different rates and with incompatible conventions. Consolidating diverse open data sources has enabled generalist models in computer vision, language, and robotics. In driving, such consolidation has so far been limited in scope to support specific sub-areas like perception, planning, or simulation. We present 123D, an open-source framework that unifies multi-modal driving data in a shared format, storing each modality at its native rate, and supporting multiple tasks via a common API and an interactive 3D viewer. We consolidate eight real-world datasets spanning about 3,400 hours and 97,000 kilometers with diverse sensor rigs and domains, and provide a CARLA simulator pipeline that can collect synthetic data at scale, demonstrated with our L3AD dataset. Across 3D object detection, reinforcement learning for planning, and end-to-end driving, we find that models trained on dataset mixtures generalize best, even at the same training budget.
Code and documentation are available at \href{https://github.com/kesai-labs/py123d}{https://github.com/kesai-labs/py123d}.
\end{abstract}

\section{Introduction}
\label{sec:introduction}

Progress in autonomous driving research is strongly tied to dataset releases. 
Most milestones in the field, from modular perception~\citep{Geiger2012CVPR, Caesar2020CVPR, Sun2020CVPR} to behavior prediction~\citep{Chang2019CVPR, Ettinger2021ICCV, Houston2021CORL} and end-to-end driving~\citep{Xu2025ARXIV, Nvidia2025PAIAV} have introduced new datasets, expanding a still-growing collection of driving recordings.

This collection is rarely studied as a whole. 
Models are typically trained and evaluated on splits of a single dataset. 
Each dataset carries biases: a particular sensor configuration, vehicle type, geographic coverage, and time period~\citep{Wang2020CVPRb, Liu2024TIV}. 
The same biases surface at deployment: sensor stacks and operating domains shift over time~\citep{Wayve2025MULTI, Waymo2025BLOG, Waymo2026BLOG, Nvidia2026HYPERION}, and re-collecting data for every deployment is impractical for driving technology meant to generalize across platforms and locations. 
We therefore consider learning driving models that exploit heterogeneous data and generalize across these biases a central open problem, and one that cannot be studied on a single dataset.

\begin{figure}
    \centering
    \includegraphics[width=\linewidth]{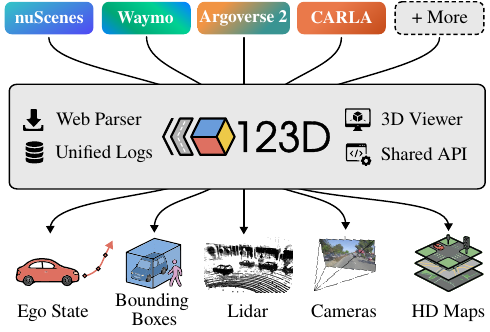}
    \vspace{-0.75cm}
    \caption{\textbf{123D.} 
    An open-source toolkit to consolidate fragmented driving data through a unified format and API for annotations, sensor modalities, and HD maps.}
    \label{fig:teaser}
    \vspace{-0.5cm}
\end{figure}

Other fields have built community-driven infrastructure for consolidating data. 
Shared corpora and libraries made training across datasets routine in language~\citep{CommonCrawl, Lhoest2021EMNLP}, and LeRobot~\citep{Cadene2024Lerobot} and Open X-Embodiment~\citep{open_x_embodiment_rt_x_2023} did the same for robotics, where the union of narrow datasets yielded positive transfer across platforms~\citep{Black2024ARXIV}.
Driving lacks an equivalent that spans all modalities: each dataset is released in its own formats, frequencies, conventions, and label sets. 
A recent position paper identifies data fragmentation as a central bottleneck of the field~\citep{Yoo2026NEURIPS}. 

In this paper, we address this fragmentation with 123D: \underline{\textit{one}} interface for \underline{\textit{2}}D and \underline{\textit{3}}D \underline{\textit{d}}riving data (\figref{fig:teaser}). 
We propose a shared format that maximizes compatibility with existing datasets: each available modality is stored as an independent, timestamped event stream without a prescribed rate, coordinate and map conventions are unified, and we preserve native labels. For this work, we consolidate eight real-world datasets and provide a pipeline that collects any amount of synthetic data in CARLA~\citep{Dosovitskiy2017CORL}. 

Using 123D, we conduct a controlled study of cross-dataset generalization: we train 3D object detectors, reinforcement-learning planners, and end-to-end driving policies at fixed data and compute budgets, and evaluate each across datasets, including domains or sensor rigs unseen during training. 
In all three tasks, models trained on dataset mixtures match or approach single-dataset models on their own data and transfer better to unseen datasets.
For example, in end-to-end driving, despite PAI-AV~\citep{Nvidia2025PAIAV} being the largest and most diverse single dataset, policies trained on 1,600 hours of  recordings from its camera rig score below 20\% on HUGSIM~\citep{Zhou2025PAMI}. Adding four smaller datasets at the same training budget raises this by 16 to 23 points.

Our contributions are: 
\textbf{(1)}~We release 123D, an open-source framework that converts diverse driving datasets into one format. It provides shared APIs for scenes and maps, an interactive 3D viewer, and a CARLA pipeline for synthetic data.
\textbf{(2)}~We build training and evaluation pipelines on this format that train on any mixture of datasets: multi-view 3D object detection, multi-agent reinforcement learning through a simulator interface, and end-to-end driving with open-loop metrics and closed-loop evaluation in multiple simulators.
\textbf{(3)}~We use these pipelines to study cross-dataset generalization in all three tasks, training on single datasets and on mixtures at a fixed budget and evaluating on datasets and simulators held out from training. 

\section{Related Work}  
\label{sec:rel_work}

\boldparagraph{Driving Datasets.} Since KITTI~\citep{Geiger2012CVPR} laid the foundations with stereo camera and lidar data, public driving datasets have grown considerably in scale and sensor coverage. Each release is typically introduced for a specific set of benchmark tasks, \eg, perception~\citep{Caesar2020CVPR, Sun2020CVPR, Liao2022PAMI}, motion forecasting~\citep{Chang2019CVPR, Wilson2021NEURIPS, Ettinger2021ICCV}, data-driven planning~\citep{Karnchanachari2024ICRA}, or end-to-end driving on either real~\citep{Xu2025ARXIV} or synthetic data~\citep{Sima2024ECCV, Jia2024NEURIPS, Nguyen2026CVPR}. Often, these data sources are shaped by task and engineering conventions of the organization that released them. Moreover, several datasets have been repurposed to new emergent tasks as the field progresses~\citep{Hu2023CVPR, Li2024CVPR, Sima2024ECCV, Dauner2024NEURIPS, Li2025ARXIV, Cao2025CORL}, demonstrating the need for tooling that handles general driving recordings uniformly. In this work, we provide such tooling capable of handling diverse datasets to support research on a wide variety of autonomous driving tasks.

\boldparagraph{Cross-Dataset Frameworks.} In robotics, shared data formats and tools~\citep{Cadene2024Lerobot, open_x_embodiment_rt_x_2023} have simplified cross-embodiment training and accelerated progress. The driving community has pursued similar efforts, but each is purpose-built for a narrow family of tasks, \eg, motion prediction and planning~\citep{Ivanovic2023NIPS, Li2023NEURIPS, Li2024TIV, Feng2024ECCV}, 3D object detection~\citep{Mmdet3d2020, Openpcdet2020}, lidar segmentation~\citep{Lentsch2026CVPR}, or scene reconstruction~\citep{Tonderski2024CVPR, Chen2025ICLR}, and retains only the task-specific modalities, typically resampled to a fixed rate (\tabref{tab:frameworks}). StandardE2E~\citep{Konev2026ARXIV} targets end-to-end driving, but hard-codes synchronization into per-frame caches, crops HD maps around the ego vehicle, and reports no trained policies or cross-dataset experiments; NCore~\citep{Nvidia2026NCORELIB} and Tri3D~\citep{Granger2024TRI3D} provide no HD maps. Prior cross-dataset studies stay within one task~\citep{Zheng2023CORL, Zhang2023CVPR, Li2025ICLR, Feng2024ECCV, Charraut2025RLC}. In contrast, 123D exposes all modalities at their native rates under a shared API, and evaluates driving models in open-loop across datasets and in closed-loop simulators.

\boldparagraph{End-to-End Driving.} End-to-end policies map sensor observations to trajectories or controls~\citep{Prakash2021CVPR, Chitta2023PAMI, Hu2023CVPR} and are evaluated in open-loop on recorded logs or in closed-loop in simulation. Since the task spans sensor data, maps, and evaluation metrics, training data and benchmarks have typically been tied to a single data source: NAVSIM~\citep{Dauner2024NEURIPS, Cao2025CORL} builds on nuPlan, Bench2Drive~\citep{Jia2024NEURIPS} and LEAD~\citep{Nguyen2026CVPR} on CARLA, WOD-E2E~\citep{Xu2025ARXIV} on Waymo data, and driving foundation models such as Alpamayo~\citep{Wang2025ARXIV} on proprietary data of a single organization. Open X-AV~\citep{Nguyen2025OXAV} pre-trains on several datasets for a single benchmark. Neural simulators such as HUGSIM~\citep{Zhou2025PAMI} and AlpaSim~\citep{Nvidia2025ALPASIM} reconstruct recorded logs for closed-loop evaluation across multiple datasets, but leave training to dataset-specific pipelines. With 123D, we provide for the first time the shared infrastructure to both train policies on dataset mixtures and to evaluate them across diverse benchmarks (\secref{sec:exp_e2e}).

\begin{table}[t]
\centering
\scriptsize
\setlength{\tabcolsep}{1.6pt}
\renewcommand{\arraystretch}{1.1}
\newcommand{\rothead}[1]{\makebox[0.9em][l]{\rotatebox{45}{#1}}}
\begin{tabular*}{\linewidth}{@{\extracolsep{\fill}} l ccccccccc l}
\toprule
\textbf{Framework}
& \rothead{Cameras} & \rothead{Lidar} & \rothead{Radar} & \rothead{3D boxes}
& \rothead{Ego state} & \rothead{Tr. lights} & \rothead{HD maps} & \rothead{Async.} & \rothead{Simulator}
& \textbf{Tasks} \\
\midrule
trajdata~\citep{Ivanovic2023NIPS}  & \xmred & \xmred & \xmred & \cmgreen & \cmgreen & \cmgreen & \cmgreen & \xmred & \cmgreen & Forecasting \\
ScenarioNet~\citep{Li2023NEURIPS}  & \xmred & \xmred & \xmred & \cmgreen & \cmgreen & \cmgreen & \cmgreen & \xmred & \cmgreen & Planning \\
MMDet3D~\citep{Mmdet3d2020}        & \cmgreen & \cmgreen & \xmred & \cmgreen & \cmpart & \xmred & \xmred & \xmred & \xmred & Detection \\
OpenPCDet~\citep{Openpcdet2020}    & \cmpart & \cmgreen & \xmred & \cmgreen & \cmpart & \xmred & \xmred & \xmred & \xmred & Detection \\
OmniRe~\citep{Chen2025ICLR}        & \cmgreen & \cmgreen & \xmred & \cmgreen & \cmpart & \xmred & \xmred & \xmred & \xmred & Reconstruction \\
OmniLiDAR~\citep{Lentsch2026CVPR}  & \xmred & \cmgreen & \xmred & \xmred & \xmred & \xmred & \xmred & \xmred & \xmred & Lidar segm. \\
StandardE2E~\citep{Konev2026ARXIV} & \cmgreen & \cmgreen & \xmred & \cmgreen & \cmgreen & \xmred & \cmgreen & \xmred & \xmred & E2E Driving\\
\midrule
\multirow{2}{*}{\textbf{123D (ours)}} & \multirow{2}{*}{\cmgreen} & \multirow{2}{*}{\cmgreen} & \multirow{2}{*}{\cmgreen} & \multirow{2}{*}{\cmgreen} & \multirow{2}{*}{\cmgreen} & \multirow{2}{*}{\cmgreen} & \multirow{2}{*}{\cmgreen} & \multirow{2}{*}{\cmgreen} & \multirow{2}{*}{\cmgreen} & \begin{tabular}[t]{@{}l@{}}Det./Plan./E2E/\\Reconstruction\end{tabular} \\
\bottomrule
\end{tabular*}
\caption{\textbf{Cross-dataset frameworks.} Modalities exposed under a shared interface, whether each modality is stored as its own stream at its native rate and can be queried at any timestamp (Async.), whether the framework connects its data to a closed-loop simulator (Simulator), and the targeted tasks. \cmpart: partial coverage, \eg poses without ego dynamics, or cameras for only some datasets.}
\label{tab:frameworks}
\vspace{-0.25cm}
\end{table}

\begin{figure*}
    \centering
    \includegraphics[width=1\linewidth]{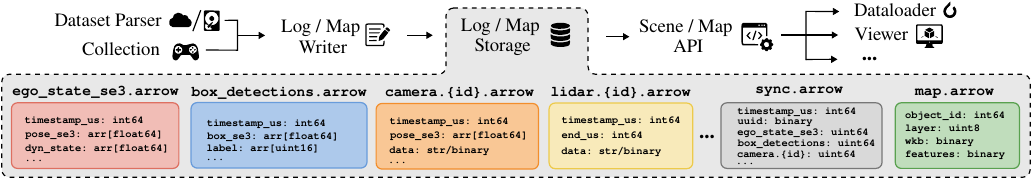}
    \vspace{-0.6cm}
    \caption{\textbf{Architecture.} Dataset parsers or simulators pass data through a shared writer to Apache Arrow~\citep{ApacheArrow} with one file per modality, a sync table, and the HD map (\secref{sec:method_one}). The Scene and Map APIs serve these files to dataloaders, viewers, and other tools (\secref{sec:method_two}).}
    \label{fig:architecture}
    \vspace{-0.25cm}
\end{figure*}

\section{The 123D Framework}  
\label{sec:py123d}

Existing driving datasets differ widely in sensor configurations, coordinate conventions, modality frequencies, map representations, and software dependencies. To address this heterogeneity, we develop \texttt{py123d}, an open-source Python package (Apache~2.0) that parses disparate sources into a uniform log and map format (\secref{sec:method_one}) and provides common data access and tooling on top of it (\secref{sec:method_two}). In the following, we describe the key challenges encountered and how our framework addresses them.

\subsection{Data Format \& Conversion}
\label{sec:method_one}

\boldparagraph{Source Formats \& Dependencies.} Each dataset comes with a different file layout, often with optional or outdated dependencies that cannot coexist in a single environment.
We encapsulate all dataset-specific logic in a \emph{dataset parser} that reads the source data, converts it to the 123D datatypes and conventions, and passes the unified log and map content to a shared \emph{log writer}/\emph{map writer} (\figref{fig:architecture}).
Each parser can read a local copy of the source dataset. Where providers permit scripted downloads (\eg,~\citep{Caesar2020CVPR, Sun2020CVPR, Nvidia2025PAIAV}), the parser can also download the source files itself for simple integration.

\boldparagraph{Heterogeneous Modalities at Varying Frequencies.} 
Recordings may be of variable durations even within a single dataset, capturing modalities at vastly different rates.
Some provide data synchronously at shared keyframes, while others are recorded asynchronously as independent events. 
To handle this inconsistency, we represent all continuous driving recordings as a \emph{log}: a directory of Apache Arrow IPC files~\citep{ApacheArrow}, with one dataframe of timestamped events per modality (\figref{fig:architecture}). 
Arrow files natively support memory mapping, which allows reading data of large modality sequences, without loading the entire episode into memory.
A separate \emph{sync table} enables cross-modal lookups by storing the matching row of every modality for each frame.
The sync table is configurable: it can preserve the source dataset's original keyframes, synchronize to a chosen reference modality (\eg, a lidar sweep), or subsample modalities to a target rate. 
Since modalities remain independent dataframes in storage, the access layer can optionally bypass the sync table and retrieve data at arbitrary timestamps. 
Moreover, static metadata (e.g., vehicle extent, sensor calibration) is embedded independently in each file's Arrow schema.
Beyond the primary modalities listed in \tabref{tab:large_dataset_comp}, 123D supports radar, semantic and instance segmentation for cameras and lidar, user-defined datatypes, and pinhole, fisheye, and f-theta camera models.

\boldparagraph{External vs. Self-contained Logs.} Raw sensor data dominates a dataset's storage footprint. 
By default, logs reference the source files \emph{externally} to avoid a second copy.
Alternatively, sensor data can be serialized into the log itself (\emph{self-contained}), which has multiple benefits. 
It avoids the small-file problem and file-count quotas of distributed storage~\citep{Aizman2019BIGDATA}, makes logs portable, and lets users choose the compression codec (JPEG, PNG, MP4 for images; Draco, LAZ, IPC for point clouds). 

\begin{figure*}[t]
     \centering
      \begin{subfigure}[b]{0.24\linewidth}
         \centering
         \includegraphics[width=\linewidth]{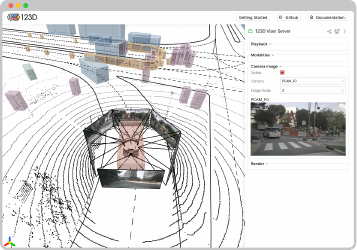}
         \caption{nuScenes~\citep{Caesar2020CVPR}}
         \label{fig:viser_nuscenes}
     \end{subfigure}
     \hfill
      \begin{subfigure}[b]{0.24\linewidth}
         \centering
         \includegraphics[width=\linewidth]{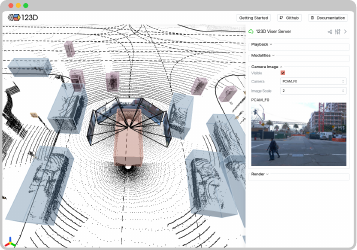}
         \caption{WOD-Perception~\citep{Sun2020CVPR}}
         \label{fig:viser_wod_perception}
     \end{subfigure}
     \hfill
     \begin{subfigure}[b]{0.24\linewidth}
         \centering
         \includegraphics[width=\linewidth]{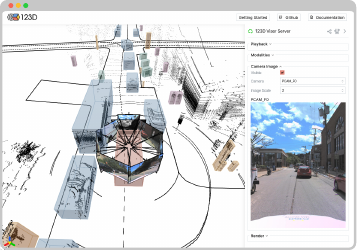}
         \caption{AV2-Sensor~\citep{Wilson2021NEURIPS}}
         \label{fig:viser_av2}
     \end{subfigure}
     \hfill
      \begin{subfigure}[b]{0.24\linewidth}
         \centering
         \includegraphics[width=\linewidth]{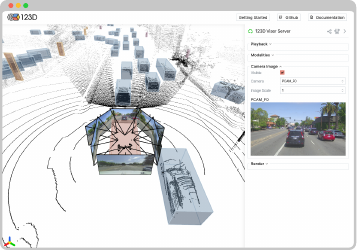}
         \caption{PandaSet~\citep{Xiao2021ITSC}}
         \label{fig:viser_pandaset}
     \end{subfigure} 

   \vspace{0.15cm} %

     \hfill
     \begin{subfigure}[b]{0.24\linewidth}
         \centering
         \includegraphics[width=\linewidth]{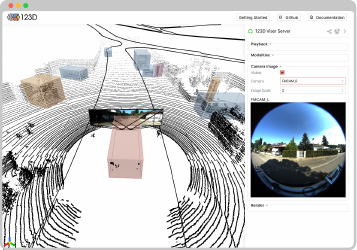}
         \caption{KITTI-360~\citep{Liao2022PAMI}}
         \label{fig:viser_kitti360}
     \end{subfigure}
     \hfill
     \begin{subfigure}[b]{0.24\linewidth}
         \centering
         \includegraphics[width=\linewidth]{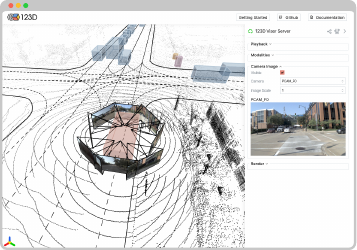}
         \caption{nuPlan~\citep{Karnchanachari2024ICRA}}
         \label{fig:viser_nuplan}
     \end{subfigure} 
     \hfill
      \begin{subfigure}[b]{0.24\linewidth}
         \centering
         \includegraphics[width=\linewidth]{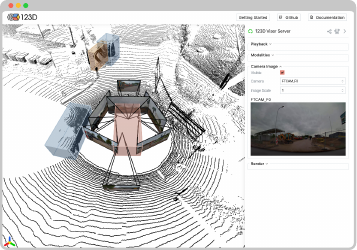}
         \caption{PAI-AV~\citep{Nvidia2025PAIAV}}
         \label{fig:viser_paiav}
     \end{subfigure}
     \hfill
     \begin{subfigure}[b]{0.24\linewidth}
         \centering
         \includegraphics[width=\linewidth]{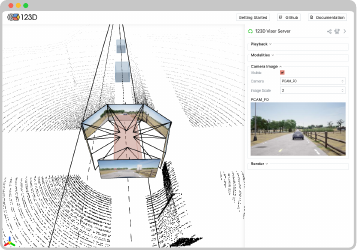}
         \caption{CARLA~\citep{Dosovitskiy2017CORL}}
         \label{fig:viser_carla}
     \end{subfigure}

     \vspace{-0.1cm}
     \caption{\textbf{3D Viewer.} Analyzing driving recordings requires frequent visual inspections. We show visualizations of supported datasets in~\ref{fig:viser_nuscenes}-\ref{fig:viser_carla} from our interactive 3D viewer based on Viser~\citep{Yi2025ARXIV}.}
     \label{fig:viser}
     \vspace{-0.25cm}
\end{figure*}

\boldparagraph{Structural Variance in HD Maps.} HD maps differ across datasets in semantic granularity, coverage (city-wide vs.\ per-log), and dimensionality (2D vs.\ 3D). 
We define a vector map that covers the superset of objects across supported datasets: \textcode{Lane} (centerline, boundaries, speed limit, neighbors), \textcode{LaneGroup} (co-directional lanes), and \textcode{Intersection} form a lane graph. Further polygon types cover surfaces such as crosswalks, carparks, and stop zones, and polylines mark road edges and road lines. 
Each map object supports 2D or 3D variants, depending on the source dataset.
If applicable, we infer missing geometries from other map elements (\eg, road edges from road polygons) with documented heuristics in the supplementary material.
All elements are stored in one Arrow IPC file, per log or shared per dataset, with geometry encoded as Well-Known Binary~\citep{OpenGIS}, enabling fast bulk reads and the Sort-Tile-Recursive (STR) tree~\citep{Leutenegger1997STR} used for map queries.

\boldparagraph{Inconsistent Labels \& Conventions.} 
Label taxonomies across datasets may not fully align due to differences in the annotation guidelines. 
123D maps the labels of every dataset to 11 common semantic categories, such as vehicle, person, and two-wheeler. Existing frameworks hard-code such a mapping during conversion~\citep{Ivanovic2023NIPS, Li2023NEURIPS}. We instead keep the original labels in every log and apply the mapping at access time. Users can therefore adapt the mapping without re-converting a dataset, and every dataset stays compatible with its own benchmarks.
Coordinate conventions also differ across datasets, requiring careful alignment when combining data from multiple sources. 
To remove this burden, we enforce standardized conventions across all converted datasets: the \emph{Body Frame} for the ego vehicle pose and bounding boxes follows ISO~8855~\citep{ISO8855} ($x$: forward, $y$: left, $z$: up); the \emph{Camera Frame} uses the OpenCV convention~\citep{opencv_library} ($x$: right, $y$: down, $z$: forward); and global coordinates follow the source dataset's native definition.

\subsection{Data Access \& Visualization}
\label{sec:method_two}

\boldparagraph{Scene API.} Common workflows with driving data involve splitting recordings into sub-sequences, sampling history and future windows, and resampling between frequencies.
Our \emph{Scene API} makes these patterns declarative.
Users specify the datasets, reference frequency, and modalities, and receive lightweight scenes that load data on demand from memory-mapped logs.
Each scene returns modalities at synchronized frames, around any timestamp (exact, nearest, forward, or backward matching), or within a time window.
This is particularly useful when modalities have different native frequencies, \eg, when pairing each lidar sweep with the nearest camera frame.

\boldparagraph{Map API.} HD maps require specialized data access, such as nearest-neighbor or intersection queries over map objects, which become expensive when applied naively to large, city-wide maps. We provide a \emph{Map API}, accessible from a scene, that serves these operations through the STR tree index built at load time (\secref{sec:method_one}), \eg, for retrieving all lanes near the ego vehicle.

\boldparagraph{Plugins \& Tools.} Working with driving data involves frequent visual inspection and recurring geometric operations. We therefore include an interactive 3D viewer based on Viser~\citep{Yi2025ARXIV} (\figref{fig:viser}) for point clouds, images, boxes, and maps, and a geometry library for transformations and camera projections shared with downstream code. Conversion templates lower the barrier to adding new datasets. 
On top of this core, we build task-specific plugins that we use throughout \secref{sec:exp}: (1) \texttt{py123d\_detection}\footnote{\url{https://github.com/kesai-labs/py123d_detection}} trains and evaluates multi-view 3D object detectors on single datasets and mixtures, (2) \texttt{py123d\_flow}\footnote{\url{https://github.com/OpenGigaflow/py123d_flow}} provides the scenes and maps of all supported datasets to the OpenGigaflow simulator~\citep{Charraut2026OpenGigaflow} for multi-agent reinforcement learning, and (3) \texttt{py123d\_garage}\footnote{\url{https://github.com/kesai-labs/py123d_garage}} trains end-to-end driving policies on any mixture of datasets and evaluates them in open-loop and across existing closed-loop benchmarks.

\begin{table*}[t]
\centering
\setlength{\tabcolsep}{9pt}
\renewcommand{\arraystretch}{1.0}
\makebox[\linewidth][c]{\resizebox{1.0\linewidth}{!}{%
\begin{tabular}{c l c | rrr | cc | ccc}
\toprule
 \multicolumn{3}{c}{\textbf{}} & \multicolumn{3}{c}{\textbf{Scale}} & \multicolumn{2}{c}{\textbf{Sensors} [\#/Hz]} & \multicolumn{3}{c}{\textbf{Annotations} [\cmgreen/Hz]} \\
\cmidrule(lr){4-6} \cmidrule(lr){7-8} \cmidrule(lr){9-11}
 & \textbf{Dataset} & \textbf{Year} & \textbf{Dur.} [h] & \textbf{Dist.} [km] & \textbf{Logs} [\#] & \textbf{Cam.} & \textbf{Lidar} & \textbf{3D Box} & \textbf{Tr. lights} & \textbf{Map} \\
\midrule
\multirow{5}{*}{\rotatebox[origin=c]{90}{\scriptsize\textbf{Manual}}}
  & nuScenes~\citep{Caesar2020CVPR}     & 2020 & 5.6   & 100.9    & 1{,}000  & 6\,/\,12 & 1\,/\,20 & \cmgreen\,/\,2  & \xmred & \cmgreen \\
  & WOD-Perception~\citep{Sun2020CVPR}        & 2020 & 6.4   & 154.0    & 1{,}150  & 5\,/\,10 & 5\,/\,10 & \cmgreen\,/\,10 & \xmred & \cmgreen \\
  & AV2-Sensor~\citep{Wilson2021NEURIPS}  & 2021 & 4.4   & 87.5     & 1{,}000  & 9\,/\,20 & 2\,/\,10 & \cmgreen\,/\,10 & \xmred & \cmgreen \\
  & PandaSet~\citep{Xiao2021ITSC}        & 2021 & 0.2   & 8.3      & 103      & 6\,/\,10 & 2\,/\,10 & \cmgreen\,/\,10 & \xmred & \xmred \\
  & KITTI-360~\citep{Liao2022PAMI}       & 2022 & 2.7   & 73.7     & 9        & 4\,/\,10 & 1\,/\,10 & \cmgreen\,/\,10 & \xmred & \cmgreen \\
\midrule
\multirow{7}{*}{\rotatebox[origin=c]{90}{\scriptsize\textbf{Auto-labeled}}}
  & WOD-Motion~\citep{Ettinger2021ICCV}                    & 2021 & 574.1 & 10{,}323.5\textsuperscript{*} & 103{,}354 & \xmred & \xmred & \cmgreen\,/\,10 & \cmgreen\,/\,10 & \cmgreen \\
\cmidrule(lr){2-11}
  & nuPlan~\citep{Karnchanachari2024ICRA} & \multirow{2}{*}{2024} & 1{,}174.3 & 17{,}808.6 & 15{,}910 & \multirow{2}{*}{8\,/\,10\textsuperscript{\dag}} & \multirow{2}{*}{5\,/\,20\textsuperscript{\dag}} & \multirow{2}{*}{\cmgreen\,/\,20} & \multirow{2}{*}{\cmgreen\,/\,20} & \multirow{2}{*}{\cmgreen} \\
  & \hspace{1em}-- mini         &      & 7.2   & 103.0      & 64        &          &          &            &                   &                   \\
\cmidrule(lr){2-11}
  & PAI-AV~\citep{Nvidia2025PAIAV}                    & 2025 & 1{,}707.0 & 69{,}265.7 & 307{,}332 & \multirow{2}{*}{7\,/\,30} & \multirow{2}{*}{1\,/\,10} & \multirow{3}{*}{\cmgreen\,/\,10} & \multirow{3}{*}{\xmred} & \multirow{2}{*}{\xmred} \\
  & \hspace{1em}-- NCore~\citep{Nvidia2026NCORE}        & 2026 & 6.3   & 167.6      & 1{,}147   &          &          &            &                   &                   \\
  & \hspace{1em}-- NuRec~\citep{Nvidia2026NUREC}        & 2026 & 12.2  & 462.9      & 2{,}202   & 7\,/\,30\textsuperscript{\ddag} & 1\,/\,10\textsuperscript{\ddag} &            &                   & \cmgreen \\
\midrule
\multirow{2}{*}{\rotatebox[origin=c]{90}{\scriptsize\textbf{Synth.}}}
  & CARLA~\citep{Dosovitskiy2017CORL}    & 2017 & \textit{var.} & \textit{var.} & \textit{var.} & \textit{var.} & \textit{var.} & \cmgreen & \cmgreen & \multirow{2}{*}{\cmgreen} \\
  & \hspace{1em}-- L3AD                              & 2026 & 7.3   & 138.7      & 789       & 6\,/\,10 & 2\,/\,10 & \cmgreen\,/\,10 & \cmgreen\,/\,10 & \\
\bottomrule
\end{tabular}
}}
\caption{\textbf{Datasets.} We compare scale (duration, driven distance, number of logs), sensor setup (number, sampling rate), and annotation availability (3D boxes, traffic-light states, HD maps).
Datasets are grouped by their labeling method (manual, auto-labeling, or synthetic from simulation). \textsuperscript{*}WOD-Motion distance is computed from the non-overlapping 20\,s training files. \textsuperscript{\dag}nuPlan sensors are released for a 120\,h subset, and fully on mini. \textsuperscript{\ddag}NuRec has real PAI-AV sensors for 1{,}502 scenes (8.3\,h), and cameras can be rendered from the reconstructions.}
\label{tab:large_dataset_comp}
\vspace{-0.25cm}
\end{table*}
\section{Experiments}
\label{sec:exp}

In this section, we use 123D to study how models for autonomous driving tasks generalize across datasets.
Throughout, we train models on a single dataset or on a mixture of datasets at a fixed budget, evaluate them on held-out splits of all datasets, and hold out some datasets from training entirely. 
We apply this protocol to three tasks: multi-view 3D object detection (\secref{sec:exp_detection}), reinforcement learning for planning in a data-driven simulator (\secref{sec:exp_gigaflow}), and end-to-end driving in open and closed loop (\secref{sec:exp_e2e}).
In each, we ask whether transfer is limited by the volume of training data or by the diversity of its sources.
We visualize our results in plots and provide tables of all experiments, as well as further studies on 3D box annotations and 3D Gaussian splatting~\citep{Kerbl2023ACM, Ren2026CVPR}, in the supplementary material.

\boldparagraph{Datasets.} Producing annotations, such as bounding boxes, is labor-intensive, and the resulting trade-off between label fidelity and scale divides existing driving datasets into three categories, as shown in \tabref{tab:large_dataset_comp}:
\textbf{(1) Manual labeling.} Trained annotators draw or review boxes from the sensor stack (often across multiple passes), providing high fidelity at considerable labor cost. Supported datasets in this category are nuScenes~\citep{Caesar2020CVPR} (2\,Hz keyframes, optionally interpolated to 10\,Hz), WOD-Perception~\citep{Sun2020CVPR}, AV2-Sensor~\citep{Wilson2021NEURIPS}, PandaSet~\citep{Xiao2021ITSC}, and KITTI-360~\citep{Liao2022PAMI}, with durations ranging from 0.2 to 6.4 hours.
\textbf{(2) Auto-labeling.} Detection, tracking, and refinement pipelines generate boxes automatically from the sensor stack, covering thousands of hours of data at reduced fidelity.
In this category, we include WOD-Motion~\citep{Ettinger2021ICCV}, nuPlan~\citep{Karnchanachari2024ICRA}, and PAI-AV~\citep{Nvidia2025PAIAV}. For PAI-AV, we additionally support two released subsets: NCore~\citep{Nvidia2026NCORE} (6.3\,h) as a compact subset, and NuRec~\citep{Nvidia2026NUREC} (12.2\,h) with HD maps and 3D reconstructions which enable closed-loop simulation.
\textbf{(3) Synthetic.} Simulators provide unlimited labeled data, inferred from the ground-truth state of the virtual driving environment.
We provide a configurable collection pipeline for CARLA~\citep{Dosovitskiy2017CORL}, based on the state-of-the-art expert policy LEAD~\citep{Nguyen2026CVPR}.
To demonstrate this collection pipeline, we publish our L3AD dataset, which includes maps and roughly 7\,h of recordings with the camera placement of nuScenes (see~\figref{fig:viser_carla}), enabling the sim-to-real transfer studies in \secref{sec:exp_detection} and \secref{sec:exp_e2e}.

\subsection{Cross-Dataset Multi-View 3D Object Detection}
\label{sec:exp_detection}

We first study how well common camera-based 3D detectors generalize across sensor rigs and operating domains. 
Our study covers two representative architectures implemented on the 123D interface, PETR~\citep{Liu2022ECCV} and BEVFormer-S~\citep{Li2022ECCV}, both with ResNet50 encoders~\citep{He2016CVPR}. 
We use single-frame variants of both to focus on transfer across sensor rigs rather than on temporal fusion.
The evaluation is restricted to the \textit{vehicle} class to avoid class ambiguities across datasets, and to ground-truth boxes within a 50\,m radius that lie inside at least one camera frustum and contain at least one lidar return.
Following common practice in this setting~\citep{Wang2023CVPR, Chang2024NEURIPS, Kuang2026CVPR}, we report a modified nuScenes-detection-score (NDS) that excludes velocity and attribute errors and keeps the nuScenes center-distance thresholds of 0.5 to 4\,m.
We train on the training splits of nuScenes (2\,Hz keyframes), WOD-Perception, AV2-Sensor, nuPlan-mini, and CARLA/L3AD individually, and on a uniform mixture of all five (\textit{Mix-D5}, where D denotes Detection). 
We evaluate on their validation splits and on PandaSet, KITTI-360 (without the $180^\circ$ fisheye cameras), and PAI-AV-NCore, which no model sees during training.
To isolate data diversity from data volume, every training configuration uses a fixed budget of 30k frames (6k per dataset for Mix-D5). To isolate volume, we additionally train on Mix-D5 with 10k and 60k frames. 
Further details are in the supplementary.

\begin{figure*}[t]
    \centering
    \includegraphics[width=1.0\linewidth]{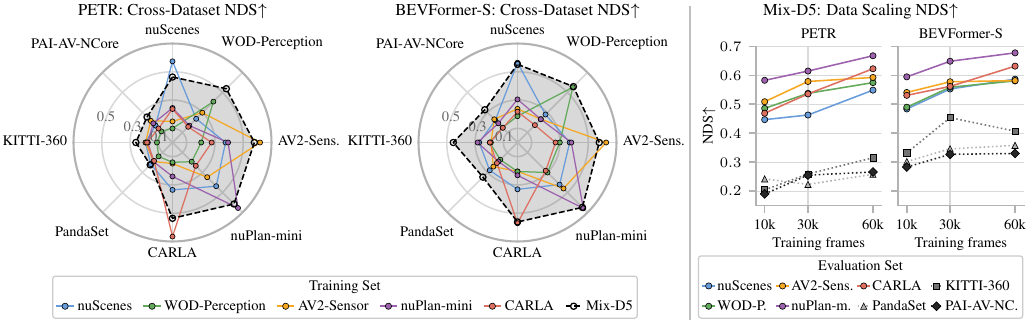}
    \vspace{-0.6cm}
    \caption{\textbf{Multi-view 3D Object Detection.} \textit{Left:} Per-dataset nuScenes detection score (NDS) for PETR~\citep{Liu2022ECCV} and BEVFormer-S~\citep{Li2022ECCV} for \textit{vehicle} detection. We evaluate on held-out validation splits of each dataset and train on \textcolor{ellisblue}{nuScenes}, \textcolor{waymogreen}{WOD-Perc.}, \textcolor{ellisorange}{AV2-Sensor}, \textcolor{tabpurple}{nuPlan-mini}, \textcolor{ellisred}{CARLA}, or a uniform mixture of these five (Mix-D5, dashed). PandaSet, KITTI-360, and PAI-AV-NCore are never seen during training. \textit{Right:} NDS of Mix-D5 trained on 10k, 30k (as on the left), or 60k frames, per evaluation dataset; held-out datasets are dotted in gray.}
    \label{fig:detection_spider}
    \vspace{-0.25cm}
\end{figure*}

\boldparagraph{Results.} The cross-dataset transfer results for both detectors are summarized in~\figref{fig:detection_spider}. 
Single-dataset models retain $0.41$–$0.67$ NDS in-domain but generally transfer poorly to other datasets ($0.10$–$0.46$ NDS). 
Transfer is asymmetric across sources: nuScenes is the most portable source relative to other single-dataset runs, while WOD-Perception is the weakest, likely due to its limited $230^{\circ}$ camera coverage seen during training. 
Despite using the nuScenes camera extrinsics, CARLA-trained models do not transfer preferentially to nuScenes (e.g., $0.24$ for PETR), similar in performance to models trained on AV2-Sensor or nuPlan-mini. 
This suggests that, for object detection, sim-to-real transfer is no easier than cross-rig transfer between real datasets, even when the simulator replicates the target rig. 

Joint Mix-D5 training closes most of these gaps, although it only trains on a fifth of each dataset's frames. It approaches or exceeds single-dataset quality on most datasets in the mixture.
BEVFormer-S benefits evenly (joint training closely matches single-dataset performance throughout), whereas PETR exhibits larger swings, such as gains on WOD-Perc. ($0.41$$\rightarrow$$0.54$) but regressions on nuScenes ($0.57$$\rightarrow$$0.46$). However, such regression largely vanishes at 60k frames ($0.55$, \figref{fig:detection_spider}, right). 

Generalization to the three fully held-out datasets (PandaSet, KITTI-360, and PAI-AV) remains the hardest task. 
BEVFormer-S is consistently better here, exceeding PETR by $0.07$–$0.20$ NDS, suggesting that explicit BEV-grid representations transfer across rigs more robustly. 
Scaling volume does not close this gap: from 10k to 60k frames, Mix-D5 improves on all five training datasets, whereas the held-out datasets gain less (PETR) or saturate at 30k (BEVFormer-S, \eg, $0.45$$\rightarrow$$0.41$ on KITTI-360). 
Cross-rig generalization thus benefits from training diversity and inductive bias, not volume alone. The remaining gaps leave cross-rig generalization as an open problem that 123D now enables studying across eight heterogeneous datasets.

\subsection{Cross-Environment Reinforcement Learning}
\label{sec:exp_gigaflow}

\begin{figure}
    \centering
    \includegraphics[width=1.0\linewidth]{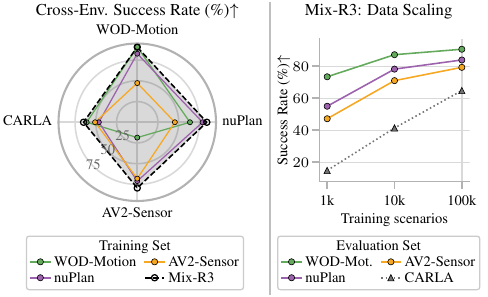}
    \vspace{-0.6cm}
    \caption{\textbf{Cross-Environment Reinforcement Learning.} \textit{Left:} Success rate ($\uparrow$) of OpenGigaflow~\citep{Charraut2026OpenGigaflow} policies trained on 100k scenarios of \textcolor{waymogreen}{WOD-Mot.}, \textcolor{tabpurple}{nuPlan}, or all three (Mix-R3, dashed), or on the 700 scenarios of \textcolor{ellisorange}{AV2-Sensor}; CARLA is never seen during training. \textit{Right:} Success rate of Mix-R3 over the number of training scenarios, per evaluation environment (CARLA dotted).}
    \label{fig:opengigaflow}
    \vspace{-0.5cm}
\end{figure}

\begin{figure*}[t]
    \centering
    \includegraphics[width=1.0\linewidth]{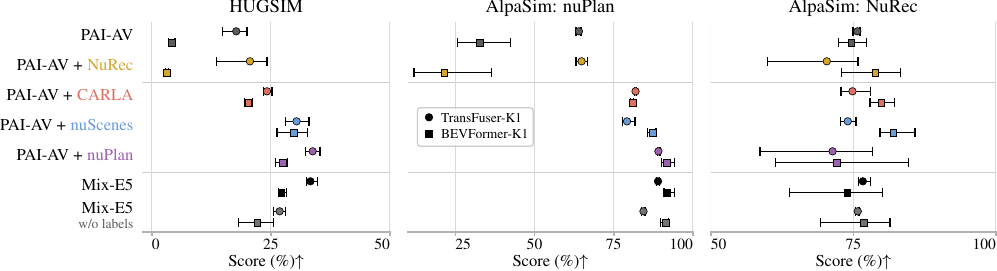}
    \vspace{-0.6cm}
    \caption{\textbf{Closed-Loop End-to-End Driving.}
    HUGSIM HD-Scores without route input, and AlpaSim scores on nuPlan (pinhole rig) and NuRec (PAI-AV rig) with route input, for TransFuser-K1 ($\bullet$) and BEVFormer-K1 ($\blacksquare$) trained on \textcolor{paigray}{PAI-AV} plus \textcolor{nurecgold}{NuRec}, \textcolor{ellisred}{CARLA}, \textcolor{ellisblue}{nuScenes}, or \textcolor{tabpurple}{nuPlan}, or all five (Mix-E5).
    Markers: mean over three seeds; bars: min and max.}
    \label{fig:e2e_closed_loop}
    \vspace{-0.25cm}
\end{figure*}

Data-driven simulators enable training and evaluation of policies with real-world maps and traffic states. 
However, they are commonly built for a single dataset~\citep{Karnchanachari2024ICRA, Gulino2023NEURIPS}. 
We demonstrate simulation across datasets by implementing a compatibility layer to OpenGigaflow~\citep{Charraut2026OpenGigaflow}, an optimized simulator for multi-agent reinforcement learning. 
We infer each vehicle's route, a necessary input for reinforcement learning planners, by matching its logged trajectory to the lane graph and extending it along the successor lanes.
We use the default policy and reward implemented in OpenGigaflow based on~\citep{Cusumano2025ICML}, which leverages self-play to control the selected actors during training.
Each actor observes a vectorized view of lanes, boundaries, traffic lights, and other actors in the scene, together with 5 goal points to reach along the extracted route. 
We train on WOD-Motion (9.1s with improved traffic lights~\citep{Yan2026Elsevier}), nuPlan (20s), AV2-Sensor (15s, no traffic lights), and mixed training of these three (Mix-R3). 
Each policy trains for 10 billion simulation steps on 100k scenarios, or on all 700 training scenarios for AV2-Sensor.
Since Mix-R3 samples uniformly from the pooled training scenarios, the overlapping WOD-Motion scenarios contribute 56.4\% of the simulation steps, nuPlan 43.5\%, and AV2-Sensor 0.1\%.
We evaluate every policy on the full validation splits of the three datasets.
As a held-out test environment, we evaluate on CARLA maps using randomly generated initial states of 40 actors and their routes in the scene, similar to~\citep{Cusumano2025ICML}. 
For all evaluations, we measure the success rate as the fraction of agents that reach their first goal point without off-road, red-light, or collision infractions.
To isolate the effect of data volume at fixed compute, we train on Mix-R3 with 1k and 10k scenarios.

\boldparagraph{Results.}
We summarize the results on held-out validation splits in~\figref{fig:opengigaflow} (left). 
Among single-environment policies, in-domain training performs best on WOD-Motion and nuPlan. 
The policy trained on AV2-Sensor transfers poorly to both, likely because its training split holds only 700 scenarios. 
The nuPlan policy, trained on 100k scenarios, even outperforms it on AV2-Sensor ($71$ vs.\ $68$\%).
Cross-training on all three real-world environments (Mix-R3) matches or exceeds the best single-source policy in each of them. 
On held-out CARLA, Mix-R3 again performs best, yet every policy stays well below its in-domain success rates.
Scaling Mix-R3 from 1k to 100k scenarios (\figref{fig:opengigaflow}, right) improves all four environments, but in-domain gains shrink beyond 10k scenarios, whereas held-out CARLA keeps improving steeply ($15$$\rightarrow$$41$$\rightarrow$$65$\%).

\subsection{Cross-Dataset End-to-End Driving}
\label{sec:exp_e2e}

\begin{figure*}[t]
    \centering
    \includegraphics[width=1.0\linewidth]{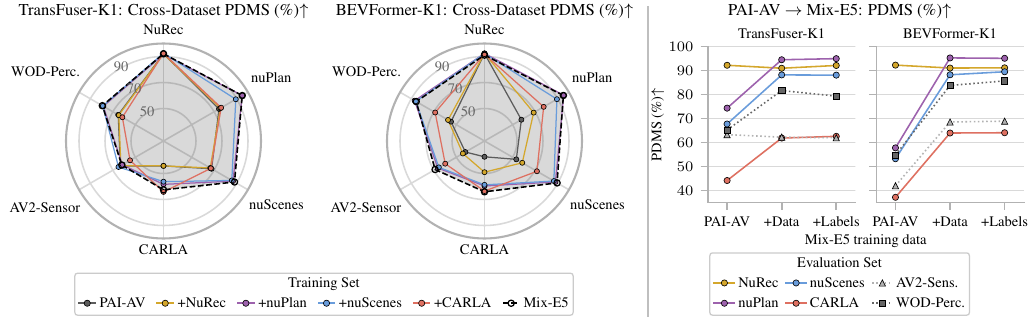}
    \vspace{-0.6cm}
    \caption{\textbf{Open-Loop End-to-End Driving.}
    \textit{Left:} PDMS of TransFuser-K1 and BEVFormer-K1 policies (mean over three seeds) trained on \textcolor{paigray}{PAI-AV} alone, on PAI-AV plus \textcolor{nurecgold}{NuRec}, \textcolor{tabpurple}{nuPlan}, \textcolor{ellisblue}{nuScenes}, or \textcolor{ellisred}{CARLA}, or on all five (Mix-E5, dashed), evaluated on held-out splits of the training datasets and on AV2-Sensor and WOD-Perc., which no mixture trains on.
    \textit{Right:} PDMS when adding the four datasets to PAI-AV with trajectory supervision only (+Data) and with their detection and segmentation labels (+Labels).}
    \label{fig:e2e_spider}
    \vspace{-0.35cm}
\end{figure*}

Finally, we study the transfer of end-to-end driving policies across camera rigs, driving styles, and regions at once.
This necessitates the unification of formats for sensors (used as inputs) as well as maps and labels (used for evaluation).

\boldparagraph{Training.}
We introduce two end-to-end driving architectures, TransFuser-K1 and BEVFormer-K1, which share their planning decoder and differ in how their encoder projects images into bird's-eye-view (BEV).
TransFuser-K1 (67M parameters) evolved from the camera-only Latent TransFuser~\citep{Chitta2023PAMI}. The BEV branch starts from learned queries instead of a lidar scan, and transformer fusion learns the mapping from image features to BEV.
BEVFormer-K1 (51M parameters) replaces this fusion with the spatial cross-attention of BEVFormer~\citep{Li2022ECCV}, which projects BEV queries into the image through each dataset's calibrated camera model.
We provide an overview of the experimental setting below, with full details in the supplementary material.

Each model receives two front-camera input frames one second apart, encoded by a ResNet-34~\citep{He2016CVPR}, the ego speed, and optionally, route target points up to $80$\,m ahead as input. A flow-matching head~\citep{Lipman2023ICLR} predicts a $6.4$\,s trajectory.
Both policies have two auxiliary heads for 3D detection~\citep{Zhou2019CenterNet} and BEV segmentation, each supervised on the datasets that provide the corresponding labels.
As end-to-end driving is challenging on small datasets alone, all policies train on ${\sim}$1,600\,h of PAI-AV (no maps) with samples 2\,s apart as the core component of their training set (2.9M samples).
To PAI-AV we optionally add the training splits of NuRec~\citep{Nvidia2026NUREC} (240k samples, real sensors, with maps), nuPlan (1.6M samples), nuScenes (83k samples), or CARLA/L3AD (140k samples, with Town13 held out), or all 5 end-to-end driving datasets (Mix-E5).
A second variant, Mix-E5 w/o labels, trains the auxiliary heads on PAI-AV's boxes only, so the added datasets contribute trajectories but no auxiliary signal from the unified representations of 123D, to isolate the benefits of the added data diversity from the added labels.
Training proceeds in three stages: perception pretraining of the encoder on the auxiliary heads, route-free midtraining of the planner, and route-conditioned posttraining with target-point dropout.
Every stage draws the same number of samples per epoch, sampled from the datasets in proportion to their sample count.
We train three seeds per configuration and report their mean.

\boldparagraph{Closed-Loop Evaluation.}
In closed-loop, the policy controls the car in simulation, and receives novel renderings from 3D reconstructions of recorded drives.
HUGSIM~\citep{Zhou2025PAMI} includes $345$ scenarios built from nuScenes, KITTI-360, PandaSet, and Waymo scenes, rendered with a nuScenes-style pinhole rig and with other traffic ranging from replayed to adversarial. We evaluate the subset of $262$ scenarios of HUGSIM after removing the $83$ scenarios that overlap with the nuScenes training split. It provides no route, and its HD-Score multiplies route completion with a per-step score of collisions, drivable area, time-to-collision, and comfort.
AlpaSim~\citep{Nvidia2025ALPASIM} provides benchmarks on $441$ NuRec validation scenes from PAI-AV (AS-NuRec) and $1{,}485$ scenes from the nuPlan test set (AS-nuPlan). It provides a route as target points $40$--$80$\,m ahead and scores each scene by the progress along the recorded drive, saturating at $80\%$, with a score of zero if the policy causes a collision, leaves the road, or leaves the corridor around the recorded path.
We report all closed-loop scores in percentages.

\boldparagraph{Open-Loop Evaluation.}
For added dataset diversity in evaluation, we also benchmark the policies in open-loop. Here, we score the predicted trajectory of each sample with our re-implementation of the Predictive Driver Model Score (PDMS)~\citep{Dauner2024NEURIPS} on 123D, which measures drivable area and route progress from the maps and at-fault collisions and time-to-collision from the box labels.
We evaluate on six datasets, including the held-out validation splits of nuPlan, nuScenes, CARLA Town13, and NuRec (which stands in for the map-less PAI-AV). The two final datasets, WOD-Perception and AV2-Sensor, are fully unseen during training, with no mixture in our experiments including them.

\boldparagraph{Results.}
\figref{fig:e2e_closed_loop} and \figref{fig:e2e_spider} summarize the closed-loop and open-loop results respectively. We draw three observations. First, mixing is an effective strategy: the best results come from combining PAI-AV with other datasets in open-loop and closed-loop evaluation.
While most gains come from adding nuPlan and nuScenes, our Mix-E5 models perform well across benchmarks with a simple mixing strategy and no additional compute cost.
Without choosing which dataset to add, Mix-E5 stays within $3$ points of the best single-dataset mixture on HUGSIM, AS-nuPlan, and all six open-loop datasets in \figref{fig:e2e_spider}.

Second, scale alone was not sufficient in our experiments: policies trained on ${\sim}$1,600 hours of PAI-AV score $92\%$ PDMS in-domain but only $37\%$–$74\%$ on the other datasets, or below $20\%$ and $65\%$ in out-of-distribution simulators like HUGSIM and AS-nuPlan. These models see only the fisheye cameras of PAI-AV. BEVFormer-K1 suffers most from this, as its projection takes the calibration as input, which is constant on a single rig.
Adding any second dataset with cameras more closely matching pinhole models raises the closed-loop scores consistently. 
For example, adding 140k CARLA samples raises BEVFormer-K1 by $16$ and $48$ points on HUGSIM and AS-nuPlan, with similar gains in open-loop (see \figref{fig:e2e_spider}). 
On AS-NuRec, which is in-domain for every mixture, adding datasets matters less: all mixtures score between $70\%$ and $82\%$. In the supplementary material, we provide comparisons to foundation models trained on significantly more proprietary data, showing that all the models we train are strong reference policies for AlpaSim. 

Third, end-to-end driving evaluation remains challenging: different benchmarks provide varying signals, and closed-loop scores vary strongly across training seeds.
The auxiliary losses in Mix-E5 show gains on HUGSIM, and for TransFuser-K1 in AS-nuPlan.
However, in open-loop evaluation and on AS-NuRec, the same labels can cause declines on average over three seeds, \eg for TransFuser-K1 on WOD-Perc. ($82\%$$\rightarrow$$79\%$ PDMS) and for BEVFormer-K1 on AS-NuRec ($-2.9$ points).
Measuring the generalization of end-to-end policies and architectures therefore requires evaluating them across many datasets and benchmarks, which is the core functionality that 123D provides.

\section{Conclusion}  
\label{sec:conclusion}

We present 123D, an open-source framework that unifies heterogeneous multi-modal driving datasets under a single interface and format.
It provides unified access to diverse sensor setups and domains, and enables model training and evaluation across several benchmarks.
Across 3D object detection, reinforcement-learning planning, and end-to-end driving, models trained on dataset mixtures generalize best, even at the same training budget.
Transfer to datasets held out from training, however, remains challenging in all three tasks.
Going further, camera-agnostic architectures and improved data mixing methods are promising avenues.
In future work, we see particular promise in curating data through our API using foundation models for reasoning~\citep{Wang2025ARXIV} and vision~\citep{Carion2026ICLR, Lin2025ARXIV}, automatically aligning annotation guidelines~\citep{Zhao2020ECCV, Zhou2022CVPR}, and filtering specific scenarios~\citep{Davidson2026CVPR}.
By releasing 123D, we lower the barrier to pursuing these directions with large-scale, diverse driving data.\looseness=-1

\boldparagraph{Limitations.} We restrict our experiments to modalities common across datasets.
123D already supports radar~\citep{Caesar2020CVPR, Nvidia2025PAIAV, Schwarzkopf2026ARXIV}, camera depth~\citep{Dosovitskiy2017CORL}, and semantic labels for images~\citep{Sun2020CVPR, Liao2022PAMI} and point clouds~\citep{Sun2020CVPR, Caesar2020CVPR, Xiao2021ITSC}, but these are available only for the subset of datasets that provide them.
Moreover, 123D unifies formats but preserves each dataset's annotations (which may differ in taxonomy, range, or visibility) and applies heuristics to infer missing map elements, which we detail in our supplementary material.
While we control for these differences in our experiments (\eg, class, range, and visibility filtering), we see high value in auto-labeling pipelines built on 123D that use foundation models~\citep{Carion2026ICLR, Lin2025ARXIV} to complete missing modalities and align annotations across datasets~\citep{Zhao2020ECCV, Zhou2022CVPR}.
The studied datasets focus on passenger vehicles, and several regions remain underrepresented.
123D is open source, and we welcome contributions that extend it to further vehicle platforms and regions.

\section*{Acknowledgments}
This work was supported by ERC Starting Grant LEGO-3D (850533) and the DFG EXC number 2064/1 - project number 390727645.
Daniel Dauner was supported by the German Federal Ministry for Economic Affairs and Energy within the project NXT GEN AI METHODS (19A23014S).
This research used compute resources at the Tübingen Machine Learning Cloud, DFG FKZ INST 37/1057-1 FUGG as well as the Training Center for Machine Learning (TCML). 
We thank the International Max Planck Research School for Intelligent Systems (IMPRS-IS) for supporting Daniel Dauner.

\section*{LLM Usage Statement}
We used LLM-based assistants (Claude) to edit the text for clarity and grammar, to help write and debug code for the framework, the training and evaluation pipelines, and the figures, and to check references and related work. All research ideas, experiments, and results are by the authors, who reviewed all text and code.

\bibliographystyle{plainnat}
\bibliography{src/bibliography_long,src/bibliography,src/bibliography_custom}

\end{document}